\documentclass[letterpaper,10pt,conference]{ieeeconf}

\IEEEoverridecommandlockouts
\usepackage{cite}
\usepackage{amsmath,amssymb,amsfonts}
\usepackage{graphicx}
\usepackage{xcolor}

\title{\LARGE \bf
Multi-material Direct Ink Writing and Embroidery for Stretchable Wearable Sensors}

\author{
Lukas Cha$^{1,2}$,
Ryman Hashem$^{1,2}$,
Ria Prakash$^{1}$,
Tanguy Declety$^{1,3}$,
Wenze Zhang$^{1}$,
and Liang He$^{1,2}$%
\thanks{$^{1}$Institute of Biomedical Engineering, University of Oxford, UK}%
\thanks{$^{2}$The Podium Institute for Sports Medicine and Technology, University of Oxford, UK}%
\thanks{$^{3}$École Polytechnique Fédérale de Lausanne, Switzerland}%
\thanks{Correspondence: \texttt{lukas.cha@eng.ox.ac.uk}}%
}

\begin{document}
\maketitle

\begin{abstract}
The development of wearable sensing systems for sports performance tracking, rehabilitation, and injury prevention has driven a growing demand for smart garments that combine comfort, durability, and accurate motion detection. This paper presents a textile-compatible fabrication workflow that integrates multi-material direct ink writing with automated embroidery to create stretchable strain sensors directly embedded onto garments. The process combines sequential multi-material printing of a silicone–carbon grease–silicone stack with automated embroidery that provides both mechanical fixation and electrical interfacing in a single step. The resulting hybrid sensor demonstrates stretchability up to 120\% strain while maintaining electrical continuity, with approximately linear behaviour up to 60\% strain (R² = 0.99), a gauge factor of 31.4, and a hysteresis of 22.9\%. Repeated loading–unloading tests over 80 cycles show baseline and peak drift of 0.135\% and 0.236\% per cycle, respectively, reflecting moderate cycle-to-cycle stability. Mechanical testing further confirms that the silicone–fabric interface remains intact under large deformation, with failure occurring in the textile rather than at the stitched boundary. As a preliminary proof of concept, the sensor was integrated into wearable elbow and knee sleeves for joint angle monitoring, showing a clear correlation between normalised resistance change ($\Delta R/R$) and bending angle. By addressing both mechanical fixation and electrical interfacing through embroidery-based integration, this approach provides a reproducible and scalable pathway for incorporating printed stretchable electronics into textile systems for motion capture and soft robotic applications.

\end{abstract}


\section{Introduction}

The integration of stretchable sensors into textiles is a rapidly advancing field with significant implications for wearable electronics, soft robotics, and human–machine interaction \cite{xiong2021functional}. In addition, these technologies play a crucial role in sports rehabilitation and performance monitoring, where real-time detection of joint movement and muscle activity can aid injury prevention and optimise athletic training \cite{de2023wearable}. Unlike rigid electronic components, stretchable devices can conform to the body’s natural motion, enabling accurate monitoring of physiological signals and joint kinematics without restricting movement \cite{atalay2018textile}. Achieving this level of mechanical compliance while maintaining reliable electrical performance remains a key challenge for the development of next-generation smart garments \cite{wang2020textile}.

Direct ink writing (DIW) has emerged as a versatile additive manufacturing technique for fabricating soft and stretchable electronics \cite{brown2025multimaterial}. It allows precise deposition of functional materials such as silicones, conductive composites, and hydrogels in custom geometries, offering excellent control over mechanical and electrical properties \cite{brown2025multimaterial, yamagishi2024direct, ho2025direct}. However, most DIW-based stretchable sensors are fabricated on standard print beds rather than directly onto textiles, requiring subsequent attachment steps to integrate the device into garments. Conventional attachment techniques such as adhesives or heat lamination often compromise the fabric’s breathability, comfort, and long-term durability \cite{sanchez2023stretchable}.

To address these limitations, textile-based fabrication strategies such as weaving, knitting, and embroidery have gained attention for their ability to embed conductive pathways and sensors directly into fabrics \cite{pu2023textile}. Among these, embroidery provides high spatial precision, pattern flexibility, and compatibility with industrial-scale manufacturing \cite{yang2025programmable}. Yet, combining embroidery with soft-material printing techniques for hybrid sensor fabrication remains relatively unexplored.

\begin{figure}[!tbp]
\centering
\includegraphics[width=\columnwidth]{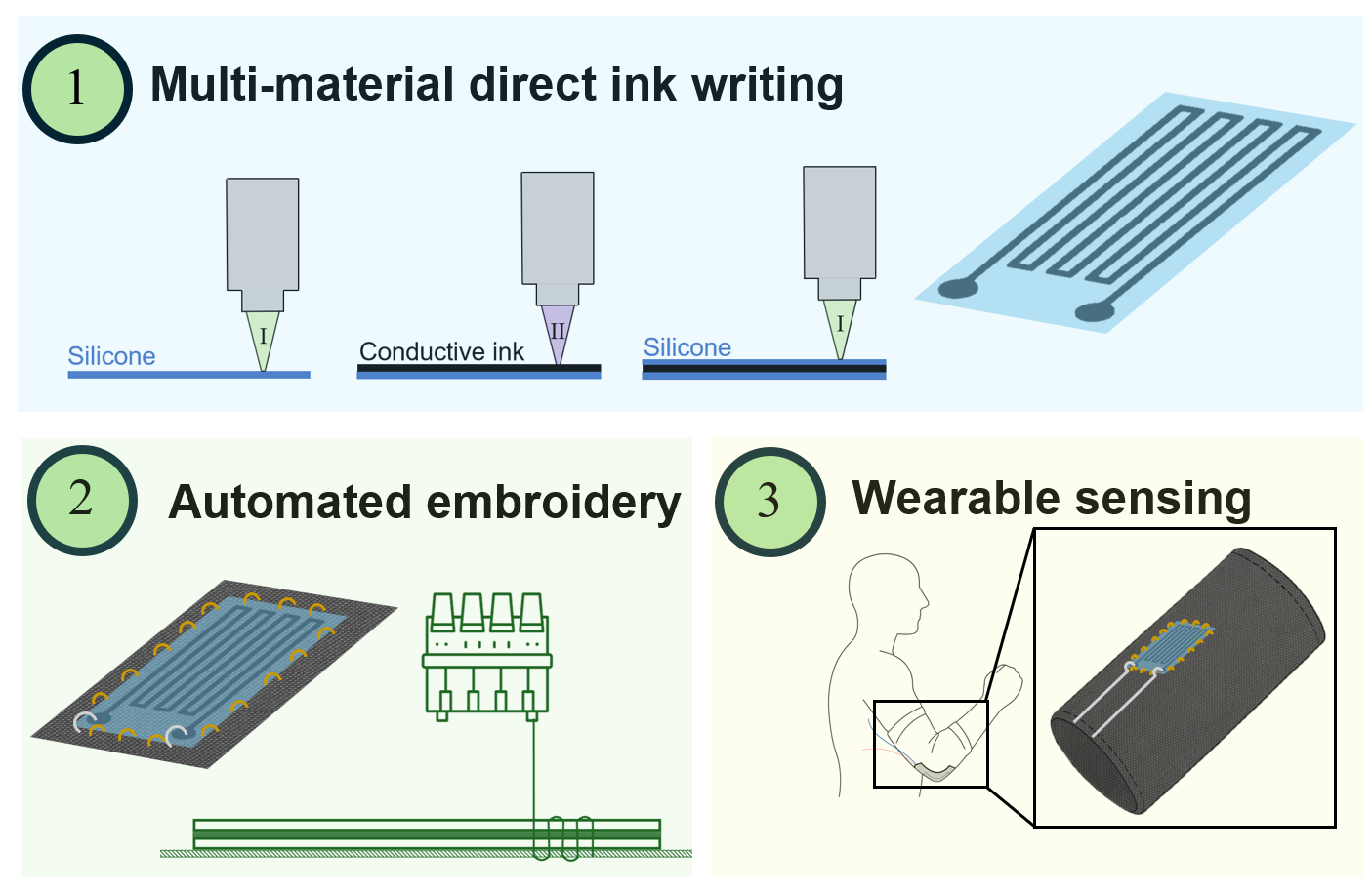}
\caption{(1) Multi-material direct ink writing of silicone and conductive ink to form a soft, stretchable strain sensor. (2) Automated embroidery mechanically anchors the printed sensor to a textile substrate while simultaneously creating electrical interconnects. (3) The integrated system enables wearable motion sensing applications.}
\label{fig:workflow}
\end{figure}

In this work, we contribute a hybrid fabrication and integration workflow for wearable movement sensing that integrates stretchable strain sensors directly onto textiles, shown in Fig. \ref{fig:workflow}. The workflow comprises:
\begin{itemize}
\item \textbf{Multi-material direct ink writing (DIW):} Printing of a resistive strain sensor using Ecoflex 00-30 as the silicone substrate and conductive carbon grease as the conductive sensing material.
\item \textbf{Embroidery-enabled textile integration:} Automated embroidery of the silicone-based sensor onto textile, simultaneously establishing mechanical anchoring (regular thread) and electrical interfacing (conductive thread) for readout of the printed sensing layer.
\end{itemize}

We demonstrate the functionality of this system through mechanical stretchability and cyclic durability tests of the resulting sensor, as well as an application experiment measuring elbow and knee joint angles. The preliminary results show that the printed–embroidered sensor maintains functional signal response under repeated strain and provides good motion detection. This method demonstrates feasibility for smart textile applications in sportswear, enabling the development of garments that enhance athletic performance through real-time motion tracking, physiological monitoring \cite{de2023wearable}, while remaining both aesthetically compatible and functionally intelligent \cite{ismar2020futuristic}.

\section{Background}

Wearable and garment-integrated strain sensors have attracted growing attention for applications in motion capture, health monitoring, and human–machine interaction. Existing textile-based sensors utilise a range of architectures—such as printed conductive films, conductive yarns, and hybrid composites—yet achieving seamless integration into garments that maintain both mechanical flexibility and stable electrical connectivity remains a significant challenge \cite{173b5b0379429fbf33662b03db4dd82abd319689, c714291ed87ffb99ee8c32d35dd893cd2630843a, e38abf3d199a16dd7daa5086c996decf831dfbca}. Below, current techniques to fabricate textile-based wearable sensors are outlined.

\textbf{Printed resistive sensors on textiles.} Screen and inkjet printing have produced high-performance strain sensors using carbon-, PEDOT:PSS-, or AgNW-based inks directly on fabric \cite{173b5b0379429fbf33662b03db4dd82abd319689,c714291ed87ffb99ee8c32d35dd893cd2630843a}. These systems deliver wide working ranges (up to 200\%) and long lifetimes (5000 cycles), with demonstrations on gloves and stretchable garments. Despite industrial maturity and good electrical performance, these methods often lack a robust electrical interfacing strategy that is mechanically stable and readily integrable with the textile substrate.

\textbf{Silicone–textile composites and encapsulated sensors.} Several studies have introduced cast or laminated elastomer layers to encapsulate conductive fabrics or inks, improving mechanical stability and stretchability on textile substrates \cite{e38abf3d199a16dd7daa5086c996decf831dfbca}. A representative work demonstrated silicone-encapsulated conductive fabric for elbow and respiration sensing with gauge factors around –1.1 and low hysteresis (3.2\%). Such composites highlight the benefit of soft encapsulation but generally rely on manual casting and external electrical connectors, rather than an automated fabrication process.

\textbf{Embroidery and stitched sensors.} Embroidery has emerged as a precise and scalable textile manufacturing method for placing conductive yarns in programmable patterns \cite{colli2023design}. Conductive thread embroidery has been used to realise resistive strain sensors with zigzag or pre-strained patterns for knee, gait, and elbow monitoring \cite{150408e453a5c4424a4590604cbac9f161f16aa2,16e1a78b6dfa3d80b6efadf10a6e27ff65a298bc}. These studies demonstrate that stitch geometry and substrate elasticity significantly affect sensitivity and strain range (up to 60–70\%). However, the embroidered thread typically functions as the sensor itself, not as an interface to a printed multilayer structure, leaving open the opportunity for hybrid printed–embroidered architectures.

\begin{figure}[!tbp]
\centering
\includegraphics[width=\columnwidth]{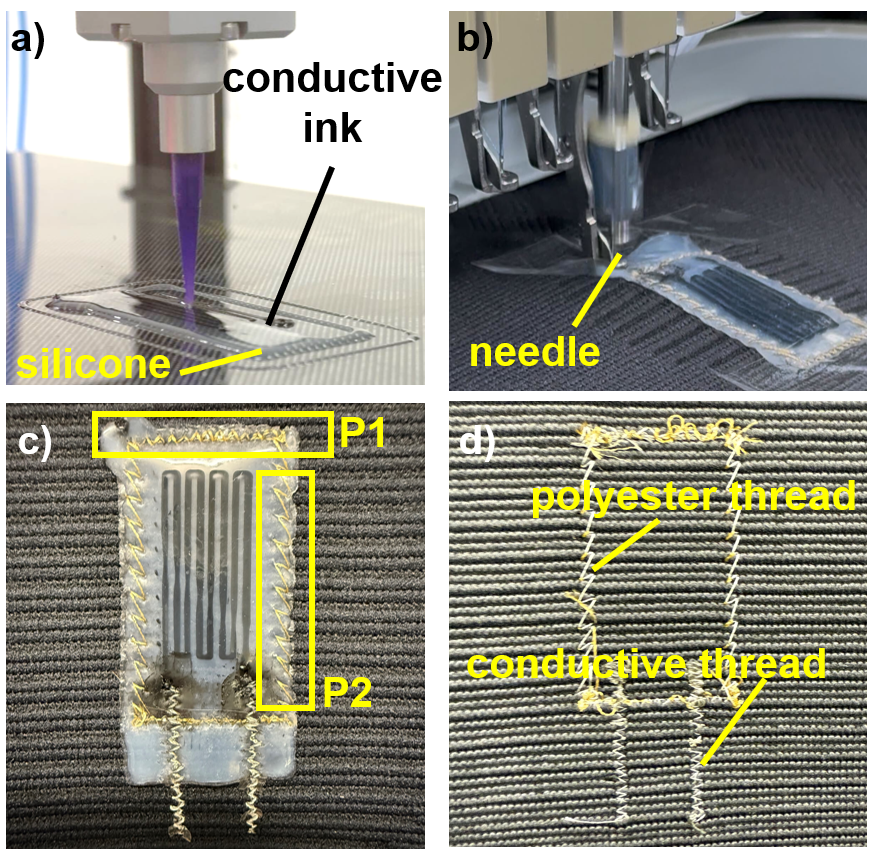}
\caption{Fabrication and integration of the printed strain sensor. (a) Direct ink writing process of the sensor (b) Embroidery step used to stitch and mechanically anchor the printed sensor to the fabric. (c) Front view of the integrated strain sensor showing the printed and stitched structure. (d) Back view illustrating the embroidered conductive interconnects.}
\label{fig:fabrication}
\end{figure}

\textbf{Hybrid printed–embroidered interconnects.} A smaller number of works explore the interface between printed conductors and embroidered threads, notably using overstitching of printed pads followed by overprinting or lamination to stabilise contact resistance under mechanical and thermal stress \cite{812822ffc9207b119fca904f4e4464964a23332b}. This approach directly addresses interconnect reliability, a common failure point in soft e-textiles, but has not yet been combined with printed elastomer–conductor stacks or demonstrated on garments for motion sensing.

\textbf{On-body validation and performance gaps.} Across the literature, printed and embroidered sensors have been validated on-body for gestures and joint-angle tracking \cite{173b5b0379429fbf33662b03db4dd82abd319689,e38abf3d199a16dd7daa5086c996decf831dfbca}, \cite{150408e453a5c4424a4590604cbac9f161f16aa2,6c2b8e90d0686760c151b435d03f889a95be96fc}. Working strains span from 0–200\%, and gauge factors vary widely depending on the conductive mechanism. Yet, explicit electrical failure criteria, cyclic drift data, and long-term embroidered–to–printed interconnect characterisation are rarely provided. Consequently, while the field demonstrates strong component technologies, a unified, garment-compatible process that combines embroidered elements with on-body evaluation remains unrealised. This gap motivates the present work, which combines printed fabrication of a strain sensor  with embroidery for simultaneous mechanical and electrical integration on fabric for wearable sensing applications.

\section{Method \& Fabrication}
The fabrication framework, depicted in Fig. \ref{fig:workflow}, is composed of a sensor printing stage and a fabric-attachment via embroidery stage. These two stages are shown in Fig. \ref{fig:fabrication}a and \ref{fig:fabrication}b, respectively. 

\begin{figure}[!tbp]
\centering
\includegraphics[width=\columnwidth]{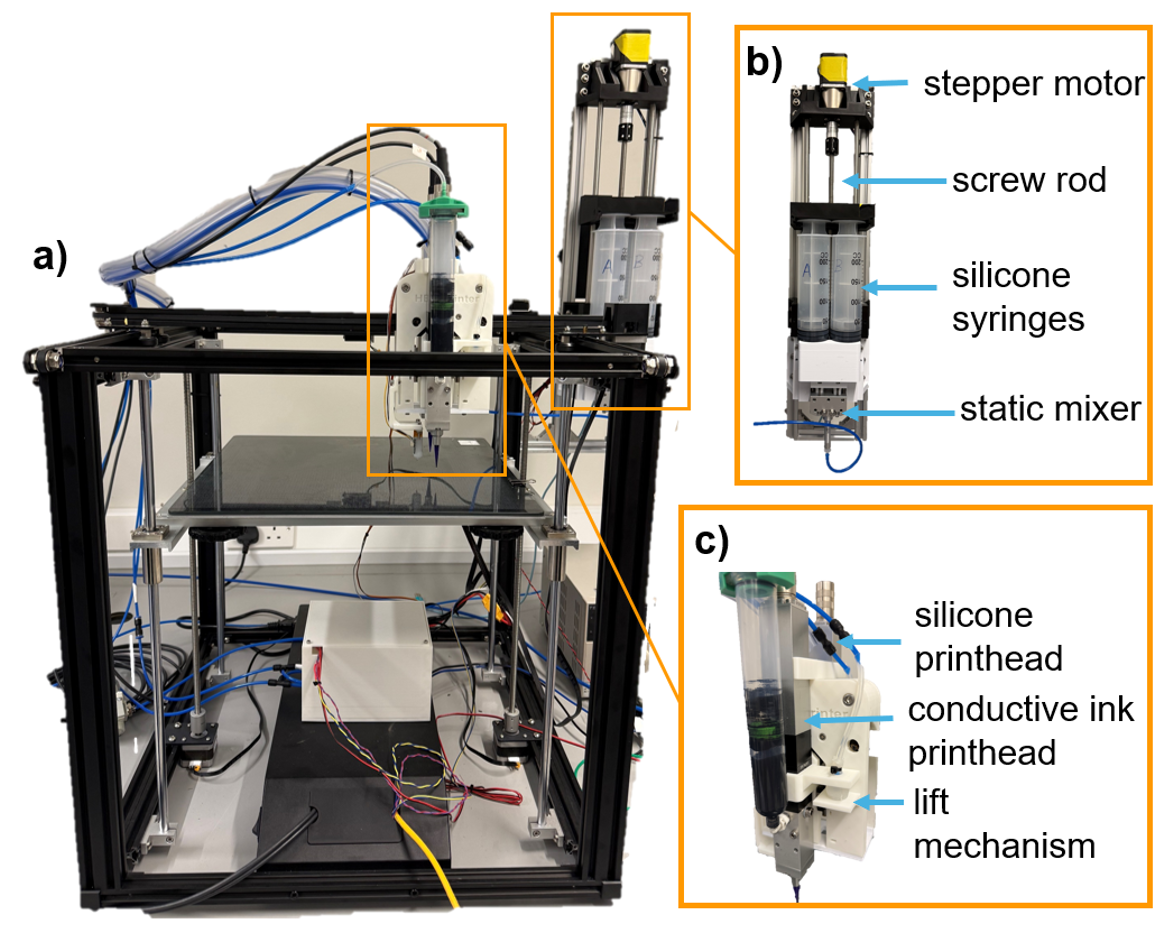}
\caption{Custom multimaterial 3D printing system. (a) Overview of the modified printer structure based on an Ender 5 Plus platform. (b) Syringe-pushing assembly for silicone mixing. (c) Dual printhead rack featuring separate printheads for silicone and conductive ink.}
\label{fig:printer}
\end{figure}

\textbf{Direct Ink Writing (DIW) of Multilayer Stretchable Sensor.}  
The stretchable strain sensor was fabricated using a modified Ender 5 Plus 3D printer equipped with two \textit{ViPro 3} (Viscotec, Germany) progressive cavity pump printheads, controlled using a Duet 3 mini+ (Duet3D, United Kingdom) board. The printheads are both equipped with a $0.51\,\text{mm}$ nozzle, resulting in the same printed line width. Each printhead was dedicated to one material: conductive carbon grease  (846-1P, MG Chemicals, Canada) and silicone (Ecoflex 00-30, Smooth-On, USA). The printheads were mounted on a custom-designed rack with linear guides actuated pneumatically as shown in Fig. \ref{fig:printer}, enabling vertical motion to lift the inactive nozzle and lower the active nozzle during printing to prevent undesired surface contact. A custom-built syringe-pushing rig continuously mixed Ecoflex 00-30 Part~A and Part~B using a static mixer, feeding the blend to the silicone printhead via a flexible tube. Ecoflex 00-30 was selected for its high elongation and compliance in the cured state. The sensor consisted of three 0.3 mm layers of silicone, carbon grease, and silicone. The print bed was maintained at 60\,$^\circ$C to accelerate partial curing of the first layer and ensure structural integrity before deposition of subsequent layers. Fig. \ref{fig:fabrication}a depicts the printing. The size of the strain sensor is $25\,\text{mm} \times 58\,\text{mm}$ with a strain gauge pattern of dimension $13\,\text{mm} \times 30\,\text{mm}$ and line width 0.51 mm. This conductive trace geometry resulted in a baseline resistance of 2.5 $M\Omega$.

\textbf{Embroidery-Based Integration onto Fabric.}
The printed sensor was integrated onto textile using a PR1055X embroidery machine (Brother, Japan), with stitching patterns designed in Hatch~3 embroidery software (Wilcom, USA). The CAD design of the sensor was imported into Hatch~3, enabling a precise placement of the embroidery stitches relative to the printed sensor. Two stitch patterns with distinct functional roles (Fig.~\ref{fig:fabrication}c) were employed:
\begin{itemize}
\item \textbf{P1: Dense line pattern.} At the top and bottom of the sensor, a dense (triple-run on Hatch 3) line stitch pattern was used to mechanically anchor the silicone sensor to the fabric, providing strong fixation. Conductive thread was also embroidered using this dense linear pattern to form electrical traces from the conductive pad region of the sensor to external contact points interfacing with a multimeter for signal acquisition.
\item \textbf{P2: Sparse zig-zag pattern.} Along the stretch-axis of the sensor, a sparse (single-run on Hatch 3) zig-zag stitch pattern (4~mm stitch separation width and 3~mm pattern width) was used to avoid restricting the natural stretchability of the underlying fabric, thereby minimising any impact on the mechanical properties of the garment.
\end{itemize}
This embroidery process achieved both mechanical anchoring and electrical interfacing, ensuring a robust yet stretchable integration of the printed sensor onto the garment substrate. The front and back sides of the resulting sensor are shown in Fig.~\ref{fig:fabrication}c and Fig.~\ref{fig:fabrication}d.

\section{Experimental Setup and Design}

\begin{figure}[t!]
\centering
\includegraphics[width=\columnwidth]{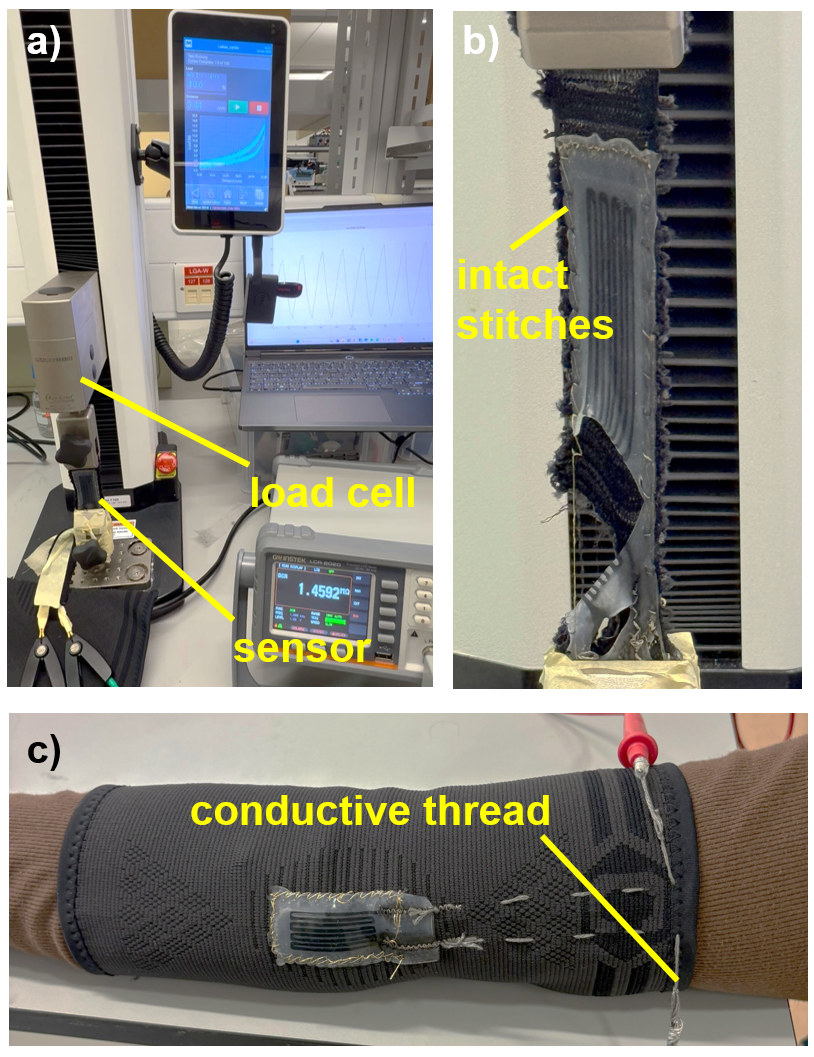}
\caption{Characterisation and demonstration of the printed strain sensor. (a) Tensile testing setup showing the sensor mounted on the universal testing machine. (b) Sensor condition after mechanical failure during the stretch-to-failure test. (c) Integration of the printed and embroidered sensor onto an elbow sleeve for wearable motion sensing.}
\label{fig:experimental_setup}
\end{figure}

\textbf{Sensor Characterisation with Tensile Tester.}  
The mechanical and electrical performance of the printed–embroidered strain sensor was characterised using a tensile tester (Mark-10, USA). The sensor was mounted on an $28\,\text{mm} \times 100\,\text{mm}$ fabric sample and electrically connected to an LCR meter (GW-Instek 6020, Taiwan) that continuously recorded resistance changes at a sampling frequency of 10 Hz. In the first test shown in Fig. \ref{fig:experimental_setup}a, cyclic tensile loading was applied to the sample to measure electrical resistance as a function of strain and corresponding force. The test was conducted over 80 loading–unloading cycles at a displacement rate of 60\,mm/min. This experiment evaluated the linearity, sensitivity, hysteresis, and drift of the fabric attached strain sensor under repeated mechanical deformation. In the second test, shown in Fig. \ref{fig:experimental_setup}b, a monotonic stretch to failure test was performed, where the sensor was elongated at a rate of 60\,mm/min until mechanical or electrical failure occurred. This detachment/break test assessed the ultimate stretchability of the printed sensor and the robustness of the embroidered mechanical and electrical interconnections between the sensor and the textile substrate. 

\textbf{Elbow and Knee Angle Application Experiment.}  
To demonstrate the sensor’s capability for motion tracking, an application experiment shown in Fig. \ref{fig:experimental_setup}c was conducted using an embroidered sensor integrated onto an elastic elbow sleeve and elastic knee sleeve. The sensor was stitched in place using the embroidery machine and connected via conductive thread to the LCR meter that recorded resistance at a frequency of 10\,Hz. Measurements were taken at distinct, controlled elbow and knee angles to establish a linear-fit calibration curve correlating normalised resistance change ($\Delta R/R$) with joint angle. The calibrated relationship was then validated during dynamic motion. A video of the elbow movement was analysed using Python and the OpenCV computer vision framework to extract ground truth elbow and knee angles frame by frame, which was compared against the sensor-estimated angles. This comparison allowed evaluation of the accuracy and temporal response of the sensor in capturing human joint kinematics.

\section{Results and Discussion}

This section presents the preliminary results of the cyclic strain and stretch-to-failure experiments, highlighting the sensor's linearity, sensitivity, hysteresis, stretchability and drift characteristics, summarised in Table \ref{table:metrics}.

\subsubsection{Hysteresis}
\begin{figure}[t!]
\centering
\includegraphics[width=1.05\columnwidth]{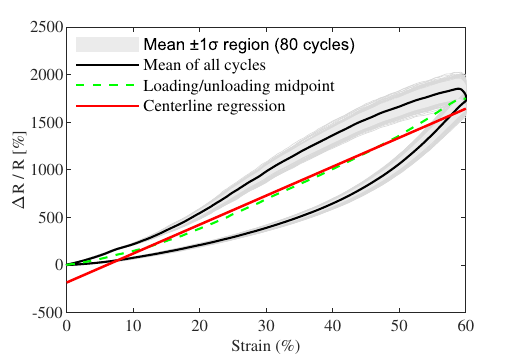} 
\caption{Cyclic strain characterisation of the printed strain sensor. The black curve represents the mean of the load and unloading sections, while the grey shaded region indicates ±1 standard deviation. The green dashed line shows the average midpoint between loading and unloading for each cycle, and the red line is the linear regression of this midpoint curve, used to evaluate sensor linearity and sensitivity.}
\label{fig:hysteresis_curve}
\end{figure}

To evaluate hysteresis behaviour, the stretch and release curves from the cyclic strain test were analysed. The relative resistance across strain is shown in Fig.\ref{fig:hysteresis_curve}. The resistive strain sensor exhibited a hysteresis of 22.9\%, calculated from the ratio of the areas under the loading and unloading curves \cite{huang2023high}. This value is consistent with other resistive strain sensors utilising carbon-based conductive networks~\cite{huang2023high}. The observed hysteresis is attributed to viscoelastic effects within the elastomer matrix and the dynamic reformation and breakdown of conductive pathways in the carbon black percolation network~\cite{huang2023high,amjadi2014highly}. Future work will investigate alternative conductive inks and matrix formulations to reduce hysteresis and enhance repeatability.

\begin{table}[h!]
\centering
\caption{Sensor Performance Metrics}
\label{table:metrics}
\begin{tabular}{lc}
\hline
\textbf{Metric} & \textbf{Value} \\
\hline
Linearity $R^2$ & 0.990 \\
Sensitivity & 31.42 \\
Hysteresis [\%]&  22.90\\
Stretchability [\%]& 120 \\
Rel. Baseline Drift/Cycle [\%] & 0.135 \\
Rel. Peak Drift/Cycle [\%] &  0.236\\
\hline
\end{tabular}
\end{table}

\subsubsection{Linearity and Sensitivity}
Linearity and sensitivity were quantified from the midpoint curve averaged over all 80 cyclic loading–unloading cycles, as shown in Fig.~\ref{fig:hysteresis_curve}. The coefficient of determination was found to be $R^2 = 0.99$, indicating highly linear behaviour. The sensitivity, defined as the slope of the linear fit, was $GF = 31.42$. These values indicate linear behaviour and gauge factor performance comparable to other silicone–carbon grease resistive strain sensors reported in the literature \cite{muth2014embedded,cha2025stretchable}.

\subsubsection{Stretchability}
\begin{figure}[t!]
\centering
\includegraphics[width=\columnwidth]{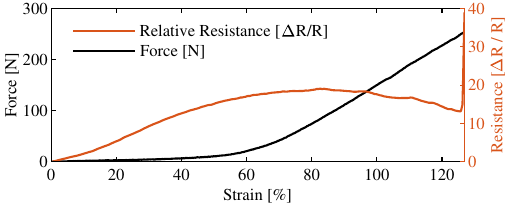} 
\caption{Stretch-to-failure characterisation of the printed strain sensor. The blue curve (left y-axis) shows the applied force as a function of strain, while the orange curve (right y-axis) represents the corresponding change in relative resistance ($\Delta$ R/R).}
\label{fig:failure_curve}
\end{figure}

The stretch-to-failure test results are shown in Fig.~\ref{fig:failure_curve}. The sensor exhibited stable and approximately linear behaviour up to 60\% strain, beyond which the response became nonlinear, and the resistance decreased slightly. This behaviour is attributed to the mechanical rearrangement of the carbon black percolation network within the conductive grease \cite{rosset2013flexible}. Mechanical failure occurred at 120\% strain, shown in Fig. \ref{fig:failure_curve}. The corresponding force curve indicates that minimal force (below 20~N) was required to stretch the sensor up to 60\% strain, indicating that it would minimally affect wearer mobility when worn.

\subsubsection{Fabric Integration}

Strain transfer between the textile and the compliant silicone layer is primarily governed by the P1 stitch regions at the top and bottom of the sensor. The dense triple-run stitching effectively fuses the silicone to the fabric along these boundaries, enabling uniform straining of both materials. The high stitch density introduces numerous attachment points, distributing load and minimising force concentration at individual perforations.

While P1 provides mechanical anchoring, the P2 stitches stabilise the sensor against out-of-plane motion. By keeping the silicone layer flat against the textile, they prevent independent flexing and promote conformal strain coupling during deformation.

Although needle perforations introduce local weak points, no visible enlargement of stitch holes or tearing was observed after stretch-to-failure testing. Notably, the fabric failed before the stitched interface, indicating that the embroidery-based anchoring remained intact. This behaviour is consistent with the lower stiffness of Ecoflex 00-30 relative to the textile, allowing the silicone to deform with the fabric without generating critical stresses at stitch locations.

\subsubsection{Cyclic Stability}
\begin{figure}[htbp]
\centering
\includegraphics[width=\columnwidth]{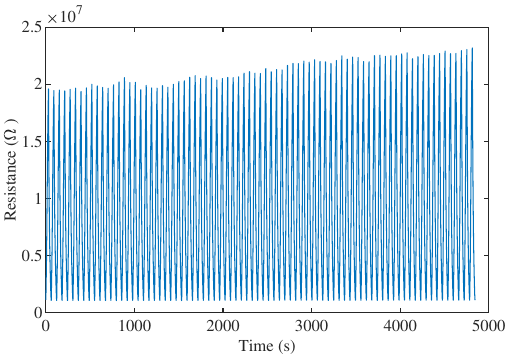} 
\caption{Absolute resistance of the printed strain sensor over 80 cyclic loading–unloading cycles. The plot shows resistance ($\Omega$) as a function of time, demonstrating consistent periodic response and stable electrical behaviour throughout repeated deformation.}
\label{fig:all_cycles_curve}
\end{figure}

The full 80-cycle test, shown in Fig.~\ref{fig:all_cycles_curve}, demonstrates gradual drift of the sensor response over time. The relative baseline drift per cycle was 0.135\%, while the relative peak drift per cycle was 0.236\% \cite{trankler2015sensortechnik}. This behaviour highlights the influence of material relaxation and conductive ink composition on long-term stability. Future work will focus on comparing different conductive printable materials to minimise drift and improve durability during repeated deformation.

\subsubsection{Elbow and Knee Motion Experiment}
\begin{figure}[t!]
\centerline{\includegraphics{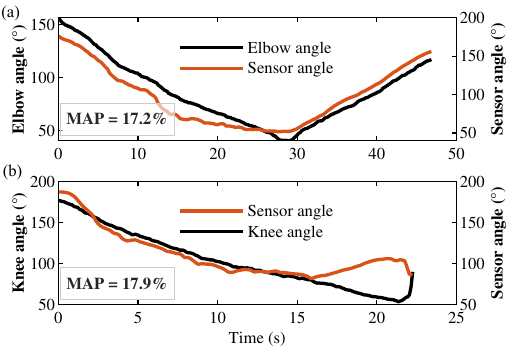}}
\caption{Calibrated joint angle tracking. (a) Elbow and (b) knee angles measured by the sensor and the OpenCV ground-truth algorithm. The mean absolute percentage error (MAPE) is shown on each plot.}
\label{fig:elbow_knee_plots}
\end{figure}

Fig.~\ref{fig:elbow_knee_plots}a and Fig.~\ref{fig:elbow_knee_plots}b show the linear-fit calibrated elbow and knee angle measurements obtained from the wearable strain sensors, benchmarked against ground-truth angle data derived from the OpenCV tracking algorithm. The mean absolute percentage error (MAPE) was 17.2\% for the elbow and 17.9\% for the knee. For the knee motion, a noticeable increase in error occurs at high flexion angles, where the sensor experiences larger strains. This behaviour is attributed to the sensor being stretched beyond its linear operating range of approximately 60\% strain, where nonlinear sensor response leads to deviation from the true angle values. In contrast, the elbow motion induces a lower overall strain on the sensor, keeping it within its linear region and thus preventing similar deviation at higher bend angles. During the experiment, the sensor exhibited noticeable temperature sensitivity, requiring constant ambient conditions for stable operation. Future work will investigate alternative conductive materials to eliminate this dependence and enhance robustness. The mean absolute error of around 17\% is higher than state-of-the-art results for garment-integrated strain sensors, e.g. \cite{atalay2018textile, 16e1a78b6dfa3d80b6efadf10a6e27ff65a298bc,0cc2218b52c61a1c3764c944aeceba1ec2b4ed98}. However, the present work primarily aims to demonstrate the fabrication framework rather than optimise sensing accuracy.

\section{Conclusion and Future Work}

This work presents a hybrid fabrication and integration strategy that combines multi-material direct ink writing of silicone and carbon grease multilayer strain sensors with embroidery for direct incorporation into textiles. The integration of printed silicone sensors into garments is often limited by challenges in mechanical attachment and reliable electrical interfacing. The proposed workflow addresses these challenges by achieving both functions within a single automated embroidery step.

The resulting resistive strain sensor demonstrates stretchability up to 120\% strain while maintaining electrical continuity, with approximately linear behaviour up to 60\% strain ($R^2 = 0.99$), a sensitivity of 31.41, and a hysteresis error of 22.9\%. Mechanical testing further indicated that the underlying fabric failed prior to the embroidered stitches, highlighting the robustness of the integration approach. Wearable demonstrations on elbow and knee joints showed clear correlation between resistance change and joint flexion, supporting the feasibility of the system for body motion monitoring. Although hysteresis, drift, and nonlinear behaviour at higher strain levels reflect limitations inherent to percolation based resistive sensing, the results establish a functional baseline for textile integrated soft sensors fabricated through a scalable workflow.

Overall, this approach contributes an integration focused solution that connects additively manufactured stretchable electronics with textile manufacturing processes.

Future work will focus on improving sensing stability and environmental robustness, including exploration of alternative conductive materials such as conductive silicones, investigation of capacitive sensing within the same framework, optimisation of stitch geometries, and development of sensor arrays capable of capturing multiple degrees of freedom in human joint motion.

\section*{Acknowledgment}

The authors would like to thank The Podium Institute for Sports Medicine and Technology, University of Oxford for funding this work. The authors would also like to thank Massimo Mariello from the Oxford Bioelectronics Lab for his help with the tensile tester.

\bibliographystyle{IEEEtran}
\bibliography{references}

\end{document}